\documentclass[sigconf]{acmart}

\AtBeginDocument{%
  }

\setcopyright{acmlicensed}
\copyrightyear{2018}
\acmYear{2018}
\acmDOI{XXXXXXX.XXXXXXX}

\acmConference[Conference acronym 'XX]{Make sure to enter the correct
  conference title from your rights confirmation emai}{June 03--05,
  2018}{Woodstock, NY}
\acmISBN{978-1-4503-XXXX-X/18/06}

\usepackage{tcolorbox}

\usepackage{algorithm}
\usepackage{algorithmic}
\usepackage{amsmath}

\usepackage{multirow}
\usepackage{booktabs}

\begin{document}

\title{Enhancing Item Tokenization for Generative Recommendation through Self-Improvement}

\author{
    Runjin Chen\textsuperscript{\rm 1, \rm 2 \dag},
    Clark Mingxuan Ju\textsuperscript{\rm 2}, 
    Ngoc Bui,\textsuperscript{\rm 2, 3 \dag}, 
    Dimosthenis Antypas\textsuperscript{\rm 2, 4 \dag}, 
    Stanley Cai\textsuperscript{\rm 2},
     Xiaopeng Wu\textsuperscript{\rm 2}, 
    Leonardo Neves\textsuperscript{\rm 2}, 
     Zhangyang Wang\textsuperscript{\rm 1}, 
   Neil Shah\textsuperscript{\rm 2},
    Tong Zhao \textsuperscript{\rm 2}
}


%


\affiliation{%
  \institution{\textsuperscript{1}The University of Texas at \city{Austin}\state{TX}\country{USA}}
  \institution{\textsuperscript{2}Snap Inc., \city{Bellevue}\state{WA}\country{USA}}
  \institution{\textsuperscript{3}Yale University, \city{New Haven}\state{CT}\country{USA}}
  \institution{\textsuperscript{4}Cardiff University, \city{Cardiff}\country{UK}}
}
\email{
  {chenrunjin, atlaswang}@utexas.edu
}
\email{{mju,scai,xwu6,lneves,nshah,tong}@snap.com}
\email{
  ngoc.bui@yale.edu, antypasd@cardiff.ac.uk
}

\thanks{\textsuperscript{\dag}Work done during an internship at Snap.}

\renewcommand{\shortauthors}{Runjin et al.}









\begin{abstract}
  Generative recommendation systems, driven by large language models (LLMs), present an innovative approach to predicting user preferences by modeling items as token sequences and generating recommendations in a generative manner. A critical challenge in this approach is the effective tokenization of items, ensuring that they are represented in a form compatible with LLMs. Current item tokenization methods include using text descriptions, numerical strings, or sequences of discrete tokens. While text-based representations integrate seamlessly with LLM tokenization, they are often too lengthy, leading to inefficiencies and complicating accurate generation. Numerical strings, while concise, lack semantic depth and fail to capture meaningful item relationships. Tokenizing items as sequences of newly defined tokens has gained traction, but it often requires external models or algorithms for token assignment. These external processes may not align with the LLM’s internal pretrained tokenization schema, leading to inconsistencies and reduced model performance. To address these limitations, we propose a self-improving item tokenization method that allows the LLM to refine its own item tokenizations during training process. Our approach starts with item tokenizations generated by any external model and periodically adjusts these tokenizations based on the LLM’s learned patterns.
  Such alignment process ensures consistency between the tokenization and the LLM’s internal understanding of the items, leading to more accurate recommendations. Furthermore, our method is simple to implement and can be integrated as a plug-and-play enhancement into existing generative recommendation systems. Experimental results on multiple datasets and using various initial tokenization strategies demonstrate the effectiveness of our method, with an average improvement of 8\% in recommendation performance.

\end{abstract}


\maketitle

\section{Introduction}
Recommendation systems~\cite{wu2024survey, li2023text} have become crucial in tailoring user experiences by accurately identifying preferences and delivering personalized content. Recently, large language models (LLMs) have made significant strides in advancing artificial intelligence across various domains, including natural language processing, information retrieval, and recommendation systems~\cite{li2023text, liu2024visual, chen2024llaga}. To harness the benefits of LLMs for recommendation tasks, researchers are now exploring a generative recommendation paradigm. This approach extends beyond traditional methods by leveraging sequence-to-sequence models as the backbone, enabling predicting user preferences in a generative manner~\cite{zhang2024generative, deldjoo2024review, ji2024genrec}, which aligns well with the capabilities of LLMs.

Generative recommendation systems operate by representing each item with a unique identifier and using an LLM to directly generate subsequent items. A central challenge in this approach lies in effectively representing items as LLM-readable identifiers, a process referred to as item tokenization. The efficiency and effectiveness of generative recommendation depend significantly on the quality of item tokenization and identification~\cite{tiger}.

Current item tokenization methods can be classified into three main types.~\cite{cid} The first is text-based tokenization, which utilizes text descriptions as item identifiers. While this approach provides rich semantic information and is naturally compatible with LLMs, the resulting prompts can become excessively long and are often challenging to generate accurately, reducing overall efficiency and effectiveness. The second is the number-based method, which uses unique numerical strings to represent items. Although shorter, numerical strings lack semantic meaning, making it challenging to convey relationships or similarities between items. The third and most commonly used method is to represent items as one or multiple Out-Of-Vocabulary (OOV) tokens. However, assigning each item a new token makes the token set excessively large, and creating challenges in learning token embeddings due to the sparsity of item interactions. 

To address these aforementioned challenges, recent research~\cite{cid, letter, tiger} focuses on representing items with multiple tokens, allowing a smaller token set to cover a larger number of items. Typically, these methods rely on an external model or algorithm, independent of the LLM backbone, to generate item identifiers. For example, TIGER~\cite{tiger} and LETTER~\cite{letter} use pre-trained text encoders (e.g., BERT~\cite{devlin2018bert}) to extract content embeddings from item descriptions, followed by RQVAE~\cite{rqvae} to construct a discrete semantic codebook that encourages items with similar descriptions to share similar token codes. Another approach, CID~\cite{cid}, applies hierarchical graph clustering on item co-occurrence matrices to assign tokens based on collaborative filtering signals.

However, item tokenization approaches that rely on external models or algorithms to generate item identifiers can create inconsistencies with the LLM backbone. The reason is that the tokenization logic often differs significantly from the one used during the LLM's pre-training. As a result, items classified as similar and assigned shared tokens by external models may not be recognized as such by the LLM. Furthermore, since item tokenization is fixed during LLM fine-tuning, the model cannot correct these errors during training. Such predefined token associations may limit the model's ability to capture item similarities accurately, potentially hindering the overall performance of the generative recommendation system. 

To address these issues, this paper proposes a \textbf{S}elf-\textbf{I}mproving method for \textbf{I}tem \textbf{T}okenization (\textbf{SIIT}) in generative recommendation. Our approach allows the LLM itself to refine item tokenization during training. Specifically, we initialize the item tokenization using any external model-generated tokens and periodically insert item and identifiers alignment tasks during LLM fine-tuning. These tasks identify outlier items that do not align well with the LLM's learned patterns, allowing their identifiers to be self-adjusted. The LLM then continues fine-tuning on the recommendation task using the updated item identifiers. SIIT is simple, easy to implement, and serves as a plug-and-play enhancement to existing item tokenization techniques. By continuously refining item tokens to better align with the LLM backbone, our approach ultimately improves the performance of generative recommendation systems.

To summarize, our main contributions are as follows: 
\begin{itemize}
    \item We analyze the shortcomings of current item tokenization methods, particularly those relying on external models or algorithms, which may not be well-aligned with the LLM. To address this, we propose a method that allows the LLM to self-tune item tokenization during training.
    \item Our approach is easy to use and serves as a plugin to enhance all tokenization methods that represent items using sequences of new tokens.
    \item Empirically, we conduct experiments on three datasets with three different item tokenization initialization methods, consistently achieving improvements over the original tokenization methods.
\end{itemize}

\begin{figure*}[t]
    \centering
    \includegraphics[width=\textwidth]{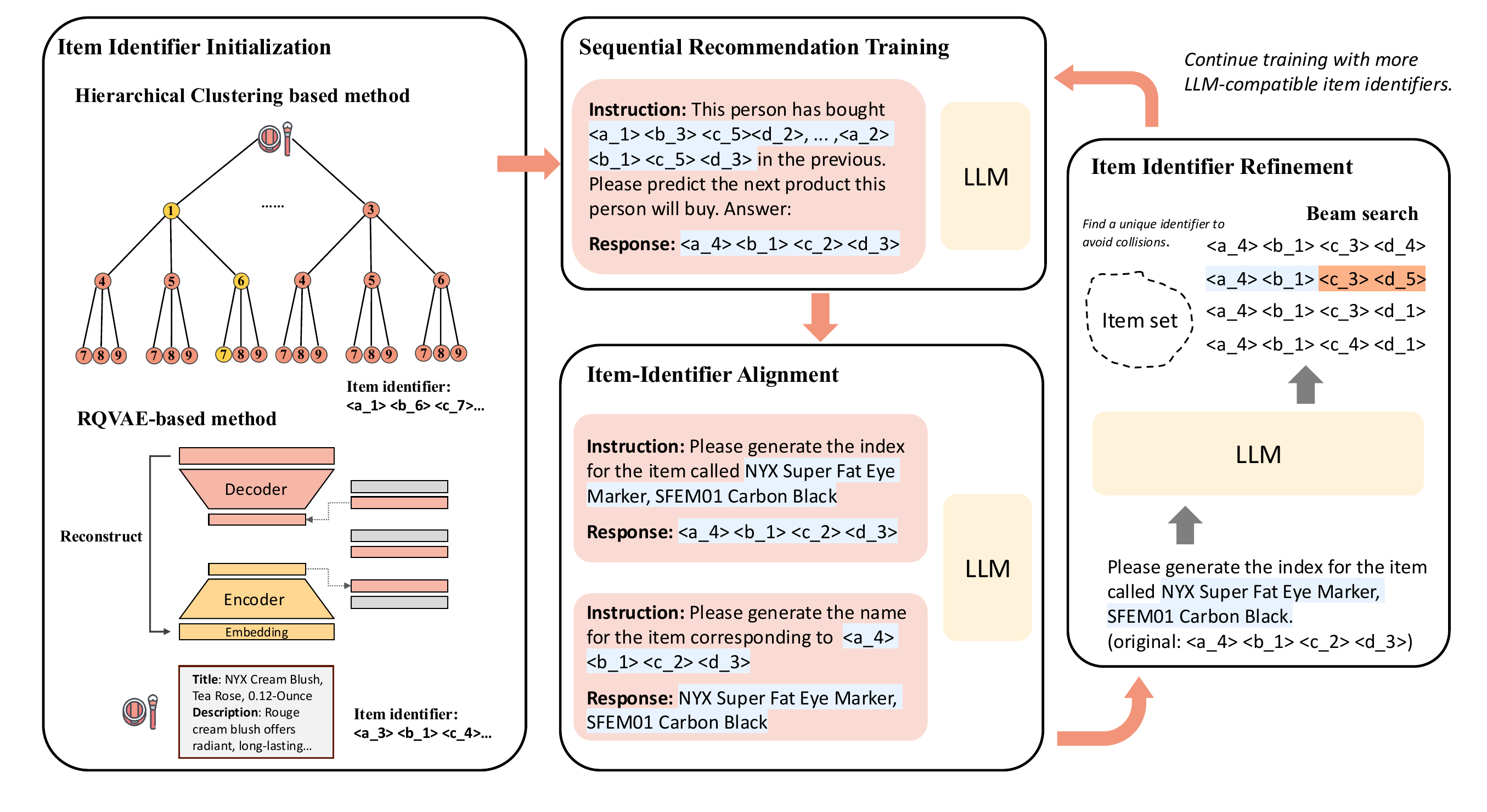}
    \vspace{-0.3in}
    \caption{The overall pipeline of SIIT. We use off-the-shelf methods to initialize item identifiers, then iteratively performs sequential recommendation training, item-identifier alignment, and identifier refinement to adjust item identifiers.}
    \label{fig:main}
\end{figure*}

\vspace{-0.3em}
\section{Preliminaries}

\subsection{Notations and Task Definition}

\noindent\textbf{Notations.} To begin, we introduce the notations used throughout the paper.  We use $\mathcal{U}$ and $\mathcal{I}$ to represent the sets of all users and items, where $u \in \mathcal{U}$ and $i \in \mathcal{I}$ denote a specific user and item, respectively. Each user $u$ has an interaction sequence $S_u = [i_1, i_2, \dots, i_k]$, and each item $i$ is associated with a semantic feature $C_i$. To integrate with LLMs, each item $i$ can be represented by an identifier that is tokenized into one or more tokens, denoted as $T_i = [t_1, t_2, \dots, t_n]$. These tokens act as unique identifiers within the LLM, and the tokenization of all items is collectively represented by $\mathcal{T}$.

\noindent\textbf{Task Definition.} Our goal is to perform sequential recommendation using LLMs. Specifically, given a user's interaction history up to timestep $k$, $S_u = [i_1, i_2, \dots, i_k]$, we aim to predict the next item the user will interact with, denoted as $i_{k+1}$.

\subsection{Item Identifier Initialization}\label{sec:init}
Our proposed SIIT can utilize any off-the-shelf item tokenization methods that employ multiple tokens as item identifiers for item identifier initialization. 
Here, we briefly introduce two popular approaches used during our experimentation: the RQVAE-based 
method and the hierarchical clustering-based method. Both serve as effective initialization techniques before our method.

\noindent\textbf{RQVAE-based Tokenization Initialization.} Given an item $i$ with its associated semantic feature $C_i$, RQVAE-based methods first obtain a semantic embedding $x_i$ using a pre-trained semantic extractor (e.g., BERT). This embedding is then compressed into a latent representation $z_i = \text{Encoder}(x_i)$ through an encoder. The latent representation $z_i$ is further quantized into a sequence of codes using a residual vector quantization process over $L$ levels, where $L$ denotes the code length. Specifically, at each quantization level $l$, a learnable codebook $E_l = \{e_k^l\}_{k=1}^N$ is used, where $e_k^l$ represents the $k$-th  codeword embedding at level $l$. The residual quantization process at level $l$ is defined as:

\begin{equation} 
t_l = \arg\min_{k} ||r_l - e_k^l||_2^2, 
\end{equation} 
\begin{equation} 
r_{l+1} = r_{l} - e_{t_l}.
\end{equation}

Here, $t_l$ represents the codeword selected at level $l$, and the residual embedding is initialized as $r_1 = z_i$. After $L$ levels of quantization, the item $i$ is represented by a sequence of codewords $T_i = [t_1, t_2, \dots, t_L]$. 

The RQVAE model is typically trained using  reconstruction loss and commitment loss. The reconstruction loss ensures that the quantized latent representation, $\hat{z}_i = \sum_{l=1}^L e_{t_l}$, can recover as much information as possible from the original semantic embedding, and is defined as $\mathcal{L}_{Recon} = ||x_i - \text{Decoder}(\hat{z}_i)||^2$. The commitment loss enforces consistency between the residual and the corresponding codebook embeddings, expressed as $\mathcal{L}_{commit} = \sum_{l=1}^L \left(||\text{sg}(r_l) - e_{t_l}^l||^2 + ||r_l - \text{sg}(e_{t_l}^l)||^2\right)$, where $\text{sg}(\cdot)$ denotes the stop-gradient operation. Additionally, some methods incorporate collaborative regularization and diversity regularization to introduce collaborative signals into the quantized codewords and enhance the diversity of codeword assignments. Further details  can be found in Appendix~\ref{app:RQVAE}.

\noindent\textbf{Hierarchical Clustering Tokenization Initialization.} Another approach to initialize item tokenization is through hierarchical clustering based on item collaborative information (i.e., CID~\cite{cid}).  CID constructs a co-occurrence matrix $O \in \mathbb{R}^{|\mathcal{I}|\times|\mathcal{I}|}$, where each entry indicates the frequency of co-occurrence between two items in a user's interaction sequence. Spectral clustering is then performed on the co-occurrence matrix $O$. If a cluster's size exceeds a predefined threshold, the clustering process is recursively applied within that cluster, yielding hierarchical clusters at increasingly finer levels, as depicted in Figure ~\ref{fig:main}. Items are tokenized according to their respective cluster center at each level, i.e., $T_i = [t_1, t_2, \dots, t_L]$, where $t_l$ denotes the cluster index at level $l$. This method encourages items with higher co-occurrence similarity to share more tokens.

\section{The Proposed Method}
Here, we outline the pipeline of our method, \textbf{SIIT}, which consists of three primary components:  fine-tuning LLMs for sequential recommendation, item-identifier alignment training, and refining item identifiers using LLMs. Each component will be discussed in detail in the subsequent subsections.
\subsection{Sequential Recommendation Task Training}\label{sec:seq}

After initializing item identifiers,  each item $i$ is represented as an identifier: $T_i = [t_1, t_2, \dots, t_L]$. To fine-tune the LLM backbone, we employ a sequential item prediction task. This involves constructing personalized recommendation prompts that incorporate the user’s historical interactions. The LLM is then tasked with predicting the next item the target user is likely to interact with, based on these instructions and the interaction history. An example prompt is shown below:

\begin{tcolorbox}[colframe=black!75!white, title=Seqential Recommendation Task]
\textbf{Instruction:} This person has bought \texttt{<a\_1> <b\_3> <c\_5> <d\_2>, ... ,<a\_2> <b\_1> <c\_5> <d\_3>} in the previous. Please predict the next product this person will buy. Answer: \\
\textbf{Response:} <a\_1> <b\_2> <c\_7> <d\_1>
\end{tcolorbox}

In our experiments, we utilize 15 templates to generate these prompts (see Appendix~\ref{app:prompt} for details). During training, the model is optimized using a next-token prediction loss applied to the generated responses. Specifically, given the instruction prompt, which includes the user’s interaction history, the objective is to generate the correct next item sequence $T_i = [t_1, t_2, \dots, t_L]$. The loss is calculated as follows:

\begin{equation}
    \mathcal{L} = -\sum_{l=1}^L \text{log } P(t_l|\text{prompt}, t_{< l})
\end{equation}


\subsection{Item-Identifier Alignment }\label{sec:align}

After a certain amount of fine-tuning on sequential recommendation tasks, we obtain the model $\mathcal{M}$. At this stage, the model has developed an understanding of the collaborative relationships between items and has integrated this knowledge into the current token embeddings.  We then periodically introduce the Item-Identifier Alignment training to establish a correspondence between the tokenized item $T_i = [t_1, t_2, \dots, t_L]$ and its original semantic feature $C_i$. This correspondence is then used to regenerate and refine the item tokenization. The Item-Identifier Alignment task consists of two subtasks: Item-to-Identifier and Identifier-to-Item. 

In the Item-to-Identifier task, we construct an instruction that includes the text feature of the item and prompt the model to generate the corresponding identifier. The prompt format is as follows:

\begin{tcolorbox}[colframe=black!75!white, title=Item-to-Identifier Task]
\textbf{Instruction:} Please generate the index for the item called \textit{NYX Super Fat Eye Marker,SFEM01 Carbon Black}:  \\
\textbf{Response:} <a\_4> <b\_1> <c\_2> <d\_3>
\end{tcolorbox}

In the Identifier-to-Item task, the instruction consists of the item identifier, and the model is expected to reconstruct the original semantic feature. The goal is to ensure that the current token embeddings of item identifier preserve the semantic feature to the greatest extent. The prompt format for this task is as follows:

\begin{tcolorbox}[colframe=black!75!white, title=Identifier-to-Item Task]
\textbf{Instruction:} Please generate the name for the item corresponding to \textit{<a\_4> <b\_1> <c\_2> <d\_3>}: \\
\textbf{Response:} NYX Super Fat Eye Marker,SFEM01 Carbon Black.
\end{tcolorbox}

For both tasks, we also apply the next token prediction loss on the model’s responses during training. After completing the training, we obtain an updated model, $\mathcal{M'}$, which will be used in the subsequent item identifier refinement process.

\subsection{Item Identifier Refinement} \label{sec:generate}



After obtaining the updated model $\mathcal{M'}$, we observed that while most items achieved a good alignment between their identifier and the corresponding item feature, a few item prompts remained challenging to optimize, consistently showing relatively high perplexity. This indicates that the current tokenization for these items does not fully match the model’s understanding. Therefore, we aim to refine the tokenization for such items based on the model’s current state. Specifically, we perform token generation for each item using the item-to-identifier task on $\mathcal{M'}$. The output reflects the identifier that the model currently finds most well-aligned with the item. 

However, this process may lead to identifier collision among similar items. Although assigning similar identifiers to items the model perceives as alike is not inherently negative, issues arise when two items share an identical identifier (identifier collision). In such cases, the model cannot differentiate between them, resulting in both items always being recommended simultaneously whenever that identifier is predicted.  Through our experiments (Section~\ref{sec: ablation}), we found that the collision can negatively impact recommendation performance. To mitigate this issue, we propose a Diverse Token Generation Strategy and a Collision Avoidance Strategy to reduce and eliminate collisions, ensuring that similar items receive different but closely related identifiers, preserving their similarity without resulting in identical identifiers.

\noindent\textbf{Diverse Identifier Generation Strategy.} To prevent item ID collisions, we use beam search instead of greedy search for generating item identifiers. For each item, beam search selects the top $k$ candidate identifiers (with $k=20$ in our experiments) that the model considers most appropriate. We then iterate through the items, checking their ranked list of generated identifiers from front to back. If an identifier has not been assigned to a previous item, it is allocated to the current item. If the identifier is already taken, we move on to the next one in the list. In the rare case where no available identifier remains after scanning the entire ranking list (an occurrence in less than 5\% of cases in our experiments), we temporarily assign the top-ranked identifier and later apply the collision avoidance strategy on them.

\noindent\textbf{Collision Avoidance Strategy.} After completing identifier reassignments for all items, there may still be rare cases where items share the same identifier due to semantic similarities in the model’s current understanding. In such cases, we append a new token to these identifiers to distinguish these items.

At this stage, each item will be assigned a newly refined, unique identifier. We then use these updated item identifiers  to continue sequential recommendation training on the previously fine-tuned model $\mathcal{M}$, which was already trained on the sequential recommendation task. After training for a few more epochs, we have the option to start the next iteration by revisiting the Item-Identifier Alignment and Item Identifier Refinement steps to further refine the tokenization.

\subsection{Inference}

After obtaining the final iteration of item identifiers and finetuned  model, we use this model and identifiers for inference. For each user in the test set, we randomly select one from 15 pre-designed sequential recommendation task prompt templates and construct an instruction based on the current user's history, which is then fed into the model. Since we need to provide a ranking list of potential next items, we apply beam search for output generation. To prevent the model from generating invalid items that are not present in the item set, we utilize constrained beam search. This is achieved by constructing a prefix tree based on all item identifiers, which constrains the output candidates at each step to only valid tokens from the prefix tree according to the current output. Finally, all outputs are ranked by their perplexity score.


In summary, we present the SIIT pipeline in Algorithm~\ref{alg:siit}. The process begins with a pretrained LLM and an off-the-shelf item tokenization method to initialize item identifiers. The model is initially fine-tuned on a sequential recommendation task using these initial identifiers. Over several iterations, the model alternates between fine-tuning on item-identifier alignment tasks, generating and refining identifiers, and fine-tuning on the recommendation task. Finally, the fully fine-tuned model and the refined identifiers are used for inference.

\begin{algorithm}[H]
    \caption{SIIT: Self-Improving Method for Item Tokenization}
    \label{alg:siit}
    \begin{algorithmic}[1]
        \REQUIRE Number of iterations $n$
        \STATE \textbf{Initialize} the LLM with pretrained model as $\mathcal{M}_{\text{init}}$
        \STATE \textbf{Initialize} item identifiers using an off-the-shelf technique as $\mathcal{T}_0$
        \STATE \textbf{Fine-tune} the model $\mathcal{M}_{\text{init}}$ on the sequential recommendation task using $\mathcal{T}_0$, yielding the updated model $\mathcal{M}_0$
        \FOR{$i = 1$ to $n$}
            \STATE \textbf{Fine-tune} the model $\mathcal{M}_{i-1}$ on the item-identifier alignment tasks using $\mathcal{T}_{i-1}$, yielding $\mathcal{M}_{i-1}'$
            \STATE \textbf{Generate} new item identifiers $\mathcal{T}_i$ using the updated model $\mathcal{M}_{i-1}'$
            \STATE \textbf{Fine-tune} the model $\mathcal{M}_{i-1}$ on the sequential recommendation task using $\mathcal{T}_i$, yielding the updated model $\mathcal{M}_i$
        \ENDFOR
        \STATE \textbf{Perform inference} on the test set using the fine-tuned model $\mathcal{M}_n$ and item identifiers $\mathcal{T}_n$ 
    \end{algorithmic}
\end{algorithm}

\section{Experiments}
We conduct a comprehensive set of experiments to validate the effectiveness of our method in various settings, addressing several key research questions:

\setlength{\leftmargini}{15pt} 
\begin{itemize} 
    \item \textbf{RQ1:} How does SIIT enhance performance compared to baseline methods in standard sequential recommendation tasks? 
    \item \textbf{RQ2:} What impact does our diverse token generation strategy and collision avoidance strategy have? 
    \item \textbf{RQ3:} How does varying the training intensity of item-identifier alignment influence item identifier refinement, and what are the effects on overall sequential recommendation task performance?
    \item \textbf{RQ4:} In what ways does SIIT enhance item identifier quality? 
\end{itemize}

\begin{table}[h]
\centering
\caption{Statistics of three public datasets after preprocessing.}\label{tab:data}
\begin{tabular}{lccc}
\toprule
\textbf{Datasets} & \textbf{\#Users} & \textbf{\#Items} & \textbf{\#Train / Val / Test} \\ 
\midrule
Instruments  & 24,772 & 9,922 & 131837 / 24772 / 24772 \\ 
Beauty  & 22,363 & 12,101 & 131413 / 22363 / 22363 \\ 
Yelp   & 30,431 & 20,033 & 225061 / 30431 / 30431 \\ 
\bottomrule
\end{tabular}
\end{table}

\subsection{Setup}
\subsubsection{\textbf{Datasets.}}
We utilize three real-world recommendation datasets from different domains, as introduced by ~\cite{letter}:
\setlength{\leftmargini}{15pt} 
\begin{itemize}
    \item \textbf{Instruments}~\cite{amazon}, sourced from the Amazon review dataset , contains user interactions with a wide range of musical equipment.
    \item \textbf{Beauty}~\cite{amazon}, also from the Amazon review dataset, features user interactions with various beauty products.
    \item \textbf{Yelp}~\cite{yelp_dataset}, a widely used dataset, includes business interactions from the Yelp platform.
\end{itemize}
We follow the preprocessing steps outlined in ~\cite{letter,lcrec}, removing users and items with fewer than five interactions. For detailed statistics, refer to Table ~\ref{tab:data}.

In the sequential recommendation task, we adopt the leave-one-out strategy as described in ~\cite{letter}. For each user, the most recent item in their interaction history is used as the test set, the second-to-last item as the validation set, and the remaining items as the training set. For each prediction sample, if a user’s interaction history exceeds 10 items, we restrict it to the 10 most recent interactions. In the text-item alignment task, we use product or place titles as textual semantic features.

\subsubsection{\textbf{Evaluation Metrics.}} We use Recall@k and NDCG@k with k=5 and 10 to evaluate model performance on sequential recommendation task. Recall@k measures the proportion of relevant items in the top-k predictions, while NDCG@k considers both the presence and ranking of these items, rewarding higher-ranked correct predictions. Our model generates the rank list from the entire item set, without predefined candidate sets.

\begin{table*}[t]
\centering
\caption{Overall performance comparison of on Instruments, Beauty and Yelp with traditional sequential baselines and LLM-based methods. R denotes Recall, and N denotes NDCG.}\label{tab:baseline}
\begin{tabular}{c|cccc|cccc|cccc}
\toprule
\multirow{2}{*}{\textbf{Methods}} & \multicolumn{4}{c|}{\textbf{Instruments}} & \multicolumn{4}{c|}{\textbf{Beauty}} & \multicolumn{4}{c}{\textbf{Yelp}} \\ 
& \textbf{R@5} & \textbf{R@10} & \textbf{N@5} & \textbf{N@10} & \textbf{R@5} & \textbf{R@10} & \textbf{N@5} & \textbf{N@10} & \textbf{R@5} & \textbf{R@10} & \textbf{N@5} & \textbf{N@10} \\
\toprule
GRU4Rec & 0.0802 & 0.1019 & 0.0666 & 0.0735 & 0.0393 & 0.0612	&0.0260	&0.0330 & 0.0183	&0.0332	&0.0114	&0.0162 \\
SASRec & 0.0792 &0.0981	&0.0673	&0.0734 & 0.0449	&0.0606	&0.0330	&0.0380 & 0.0160	& 0.0283	&0.0101	&0.0140 \\
Caser & 0.0721	&0.0917	&0.0580	&0.0643 & 0.0326	&0.0503	&0.0214	&0.0271 & 0.0184	&0.0317	&0.0113	&0.0156 \\
Bert4Rec & 0.0747	&0.0947	&0.0606	&0.0670 & 0.0404	&0.0587	&0.0282	&0.0341 & 0.0179	&0.0315	&0.0111	&0.0155 \\
MStein & 0.0756 & 0.0889 & 0.0648 & 0.0691 & 0.0466 & 0.0631 & 0.0337 & 0.0390 & 0.0146 & 0.0230 & 0.0093 & 0.0120 \\
\midrule
SID &0.0724	&0.0819	&0.0659	&0.0690 & 0.0316	&0.0449	&0.0224	&0.0267 & 0.0217	&0.0329	&0.0148	&0.0184 \\
TID & 0.0420	&0.0593	&0.0316	&0.0372 & 0.0168	&0.0334	&0.0660	&0.0834 & 0.0163	&0.0229	&0.0117	&0.0138 \\
IID & 0.0686	&0.0808	&0.0605	&0.0644 & 0.0309	&0.0456	&0.0216	&0.0263 & 0.0154	&0.0262	&0.0101	&0.0136 \\
\midrule
CID & 0.0827	&0.0998	&0.0720	&0.0775 & 0.0389	&0.0573	&0.0147	&0.0294 & 0.0272	&0.0448	&0.0175	& 0.0231 \\
\textbf{CID+SIIT} & \textbf{0.0844}	&\textbf{0.1012}	&\textbf{0.0733}	&\textbf{0.0787} & \textbf{0.0398}	&\textbf{0.0596}	&\textbf{0.0273}	&\textbf{0.0336} &\textbf{0.0290}	&\textbf{0.0468}	&\textbf{0.0194}	&\textbf{0.0251} \\
\midrule
TIGER & 0.0874	&0.1077	&0.0745	&0.0810 & 0.0426	&0.0647	&0.0287	&0.0358 & 0.0250	&0.0396	&0.0161	&0.0208 \\
\textbf{TIGER+SIIT} & \textbf{0.0916}	&\textbf{0.1127}	&\textbf{0.0779}	&\textbf{0.0847} & \textbf{0.0461}	&\textbf{0.0699}	&\textbf{0.0310}	& \textbf{0.0387} &\textbf{0.0271}	&\textbf{0.0427}	&\textbf{0.0177}	&\textbf{0.0227} \\
\midrule
LETTER & 0.0878	&0.1106	&0.0743	&0.0817 & 0.0504	&0.0770	&0.0349	&0.0435 &0.0257	&0.0408	&0.0171	&0.0219 \\
\textbf{LETTER+SIIT} & \textbf{0.0941}	&\textbf{0.1157}	&\textbf{0.0797}	&\textbf{0.0867} & \textbf{0.0543}	&\textbf{0.0790}	&\textbf{0.0367} &	\textbf{0.0446} & \textbf{0.0267}	&\textbf{0.0435}	&\textbf{0.0173}	&\textbf{0.0227} \\

\bottomrule
\end{tabular}
\end{table*}

\subsubsection{\textbf{Baselines.}} 
We evaluate our method against three categories of state-of-the-art models, offering a comprehensive comparison across diverse approaches.

\setlength{\leftmargini}{5pt} 
\begin{itemize}
    \item \textbf{Traditional Sequential Recommendation Models:}  
    \textbf{Caser}\cite{caser} employs convolutional neural networks to capture both general preferences and sequential patterns in user behavior.  \textbf{GRU4Rec}\cite{GRU4Rec} is originally designed for session-based recommendation, using GRUs to model the sequence of user interactions. \textbf{SASRec}~\cite{sasrec} leverages self-attention mechanisms to capture long-term dependencies in user interaction histories. \textbf{BERT4Rec}~\cite{bert4rec} adopts a BERT-style bidirectional language model to learn item representations. Mstein~\cite{metein} propose a  self-supervised learning framework based on the Mutual WasserStein discrepancy minimization for the sequential recommendation.
    
    \item \textbf{LLM-Based Recommendation with Simple Item Tokenization:}  
    This category consists of models that use large language models (LLMs) as the recommendation backbone, generating the next item in a generative manner with simple item tokenization methods. \textbf{TID}~\cite{cid} utilizes the item title as its identifier. \textbf{SID}~\cite{cid} assigns each item a sequential numerical ID, starting from "1001," based on its order in the user interaction history, thereby capturing co-occurrence patterns. \textbf{IID}~\cite{cid} employs a unique out-of-vocabulary (OOV) token as an identifier for each item.    
    \item \textbf{LLM-Based Recommendation with Advanced Item Tokenization:}  
    These models utilize more sophisticated item tokenization methods. \textbf{CID}~\cite{cid} applies hierarchical spectral clustering on the item co-occurrence matrix to generate item identifiers. \textbf{TIGER}~\cite{tiger} employs a codebook-based approach using RQ-VAE, quantizing item embeddings into code sequences. \textbf{LETTER}~\cite{letter} similarly utilizes RQ-VAE for generating item identifiers, incorporating collaborative and diversity regularization during training. These approaches initialize items with new token sequences, also serving as item identifier initialization method within our method.
    
\end{itemize}

\subsubsection{\textbf{Implementation Details.}}
In our experiments, we use Flan-T5-Small \footnote{\url{https://huggingface.co/google/flan-t5-small}} as the large language model backbone. The learning rate is set to 1e-3, and we employ the AdamW optimizer. The batch size is set to 256, and the warm-up period for training the sequential recommendation task is set to 20 epochs. For each iteration, we train for an additional 10 epochs for recommendation task. The number of iterations is selected from a range of 2 to 5, based on performance on the validation set. We also tune the hyperparameters for all baseline models using the validation set. All experiments are conducted on 8 NVIDIA A100-SXM4-40G GPUs.

\subsection{Compare with Baselines (RQ1)}
We evaluate SIIT on three datasets, initializing item identifiers using three different methods: CID, TIGER, and LETTER. Additionally, we compare SIIT with other LLM-based baselines as well as traditional sequential recommendation baselines. Our key observations are as follows:

First, among all LLM-based baselines, using multiple newly introduced tokens as item identifiers (e.g., CID, TIGER, LETTER) generally outperforms approaches that rely on a single token (IID), text descriptions (TID), or numerical strings (SID). While TID aligns naturally with LLMs due to its text-based nature, its use often introduces redundancy since lengthy text descriptions can dilute performance. SID, among the three basic tokenization methods, performs relatively well due to its brevity and its ability to leverage numerical tokens learned during pretraining, thus avoiding the need for learning from scratch. However, SID lacks semantic information, limiting its ability to capture item relationships. IID, which assigns a unique ID to each item, introduces many new tokens, but interaction sparsity hinders effective token learning, leading to suboptimal performance. Using multiple tokens for item identifiers resolves these issues by enabling similar items to share tokens while maintaining the uniqueness of each item’s identifier.

Second, among the multi-token methods, LETTER typically performs best across most cases, as it incorporates both collaborative and semantic information. However, on the Yelp dataset, CID outperforms the others. This can be attributed to the simplicity of Yelp’s text descriptions and the inability of RQVAE-based methods to fully harness semantic meaning, and it prevents the generation of LLM-compatible codes. As a result, the purely collaborative signal of CID becomes more effective in this case.

Most importantly, \textbf{{regardless of which initialization method is used for item identifiers, SIIT consistently improves upon the baseline methods}}. This is achieved through the identifier refinement process during training, which allows SIIT to better integrate semantic information into the identifiers using the LLM itself, making the identifiers more compatible with the semantic meaning learned by the LLM.

\subsection{Influence of Diverse Identifier Generation Strategy and Collision Avoidance Atrategy (RQ2)} \label{sec: ablation}

\begin{table}[t]
\centering
\renewcommand{\arraystretch}{1.1}
\caption{Ablation study of  diverse identifier generation strategy and collision avoidance strategy. Experiments are done on Yelp dataset} \label{tab:greedy}
\begin{tabular}{c|c|c|cc}
\toprule
 \textbf{Init.} & \textbf{Strategy} & \textbf{Collision Rate} & \textbf{R@10} & \textbf{N@10} \\
\toprule

 \multirow{4}{*}{CID}& Greedy	&23.59\%	&0.0289 &	0.0169\\
 & Greedy+CA	&0\%	&0.0380 &	0.0212\\
  & Beam & 2.73\%	&0.0420	& 0.0234 \\
  & \textbf{Beam+CA(our)} &  \textbf{0\%}	& \textbf{0.0468} &	 \textbf{0.0251} \\
  \midrule
  \multirow{4}{*}{TIGER} & Greedy & 23.97\%	&0.0258	&0.0143\\
   & Greedy+CA & 0\%	&0.0363	&0.0196\\
  & Beam & 2.63\%	&0.0371 &	0.0200 \\
   &  \textbf{Beam+CA(our)} &  \textbf{0\%}	& \textbf{0.0427}	& \textbf{0.0227} \\
  \midrule
  \multirow{4}{*}{LETTER} & Greedy & 23.58\%	&0.0292 &	0.0158\\
  &Greedy+CA & 0\%	&0.0328	&0.0173\\
   & Beam & 3.98\%	&0.0406	& 0.0217\\
   &  \textbf{Beam+CA(our)} &  \textbf{0\%}	& \textbf{0.0435}	& \textbf{0.0227}\\

\bottomrule
\end{tabular}
\end{table}

\begin{figure*}[t]
    \centering
    \includegraphics[width=\textwidth]{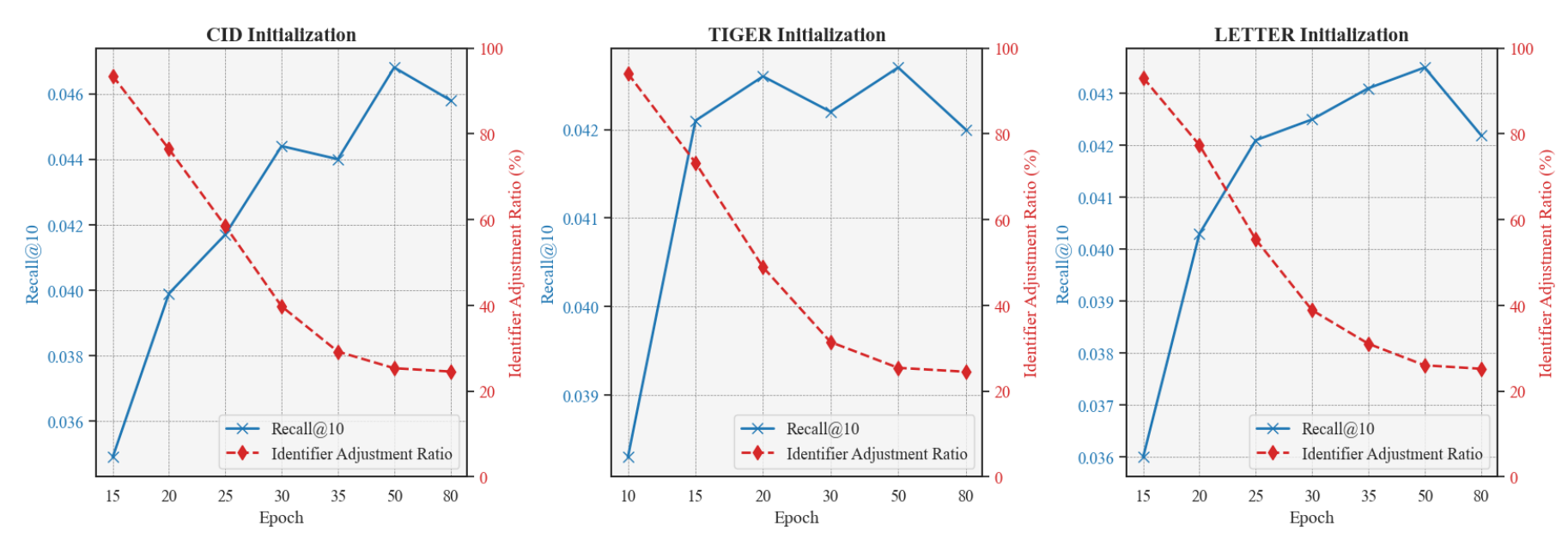}
    \caption{Influence of intensity of Item-Identifier Alignment}
    \label{fig:de}
\end{figure*}
To assess the impact of our diverse identifier generation and collision avoidance strategies, we conducted an ablation study. The diverse identifier generation strategy was removed by replacing beam search with greedy search during identifier generation. For the removal of the collision avoidance (CA) strategy, we did not add new tokens to differentiate collided items during the identifier generation process. In this scenario, multiple items could share the same identifier. When the model predicts this identifier in the recommendation task, all items associated with the identifier are included in the rank list, with their internal order shuffled.

We evaluated the recommendation performance on the Yelp dataset using identifiers generated with and without these strategies, as shown in Table ~\ref{tab:greedy}. The "Collision Rate" represents the percentage of items assigned collided identifiers in the first iteration. "R@10" and "N@10" refer to the Recall@10 and NDCG@10, respectively, for the recommendation task.

The results demonstrate that the diverse identifier generation strategy significantly reduces the collision rate, and the addition of the collision avoidance strategy eliminates it entirely. Moreover, both strategies lead to notable improvements in the downstream recommendation task performance.

\subsection{Influence of intensity of Item-Identifier Alignment(RQ3)}
We employ item-identifier alignment training to help our model establish an alignment between item semantic features and their identifiers. During this process, we apply the item-to-identifier task to identify outlier items that cannot achieve a well-aligned match with their original identifiers. When the model struggles to generate these identifiers accurately, it may indicate that the identifiers are not well-suited for the items. We leverage this characteristic to refine those identifiers. Notably, the extent of alignment can be adjusted through the training intensity (i.e. epoch), allowing us to control the degree of refinement.

During the alignment training, the model first adapts to the samples that are easiest to learn, indicating that these items' identifiers are relatively well-suited. As training progresses, the model gradually aligns with more challenging samples, eventually leaving a subset of samples that  can not align well. This process is illustrated in Figure ~\ref{fig:de}, where we vary training epochs for item-identifier alignment on the Yelp datasets. We define the \textit{"identifier adjustment ratio"} as the proportion of items with identifiers differing from their original ones in the item-to-identifier task. This ratio decreases gradually, stabilizing at a certain level. By adjusting training intensity, we can control the proportion of modified, non-compatible identifiers. As illustrated, sequential recommendation task results are visualized based on the modified identifiers. Notably, performance is suboptimal at very low training intensities, which retain only a small proportion of the most suitable identifiers unchanged. The optimal results generally occur when the identifier adjustment ratio converges to a stable level.

\subsection{Understanding how SIIT improve item Identifiers(RQ4)}
To understand how and why our identifier refinement can improve the recommendation task, we delve into the generated identifiers and conduct an analysis. We find that our refined identifiers can better capture the semantic information of items. We use a metric named \textit{semantic identifier similarity}, which evaluates the average token overlap among similar items. This is calculated as follows:

First, we calculate the token overlap matrix $O\in\mathbb{R}^{|I|\times|I|}$,where each entry $O[i][j]$ represents the number of token overlaps between the identifiers of items $i$ and $j$. Next, we calculate a semantic similarity matrix $S\in\mathbb{R}^{|I|\times|I|}$, where each entry $S[i][j]$ represents the cosine similarity between the encoded semantic features of items $i$ and $j$. We then use $S$ as the weight to calculate the weighted average of $O$ as the final \textit{semantic identifier similarity}, which demonstrates the overall similarity among identifiers of items with similar semantic meanings.
\begin{table}[t]
\centering
\renewcommand{\arraystretch}{1.1}
\caption{Semantic Identifier Similarity on Yelp dataset} \label{tab:sim}
\begin{tabular}{c|c|c|c}
\toprule
 \textbf{Init.} & CID & TIGER & LETTER \\
\toprule
Original & 0.3734 & 0.3856 & 0.0261 \\
Refined & 0.3863 & 0.3877 & 0.0302 \\
\bottomrule
\end{tabular}
\end{table}

The statistics of the original identifiers and our refined identifiers on the Yelp dataset are shown in Table~\ref{tab:sim}. We observe that our refined identifiers consistently exhibit higher semantic identifier similarity than the original ones.


\section{Related Work}
\subsection{Sequential Recommendation}
Sequential recommendation leverage a user's interaction history to predict the next item they may engage with~\cite{xie2022contrastive, chen2018sequential,chang2021sequential}. The model needs to identify underlying user interests from the sequence of past interactions and utilize collaborative signals learned from training data to make accurate predictions. Early research in sequential recommendation focused on methods like sequential pattern mining~\cite{yap2012effective, fournier2017survey} and Markov chain models~\cite{he2016fusing, rendle2010factorizing} to capture the transitions between user-item interactions.In recent years, deep learning models have emerged as more effective tools for extracting latent representations from interaction histories and searching the item space for relevant recommendations~\cite{fang2020deep, zhao2020deep}. Typically, deep-learning based sequential recommendation models represent each item as an embedding and summarize user interactions to inform predictions. For instance, GRU4Rec~\cite{GRU4Rec} employs a GRU to incorporate all interactions into a hidden state, while Caser~\cite{caser} introduces CNNs to gradually summarize item sequences. Additionally, models like BERT4Rec~\cite{bert4rec} and SASRec~\cite{sasrec} utilize self-attention and masked self-attention mechanisms to capture item relationships within the interaction sequence.However, these methods predominantly focus on collaborative signals and sequential patterns, often overlooking item content. To address this limitation, more recent approaches like FDSA~\cite{fdsa} and S$^{3}$Rec~\cite{s3rec} incorporate item content information, enriching the model's understanding of item attributes and improving overall recommendation accuracy.

\subsection{LLMs for Recommendation}
As the impact of large language models (LLMs) continues to grow across various fields, more researchers are adapting LLMs for recommendation systems~\cite{lin2024data, li2023prompt,acharya2023llm}. Current research can be categorized into two main approaches. The first approach leverages LLMs' inherent capabilities in processing text to perform recommendation tasks~\cite{hou2024large, he2023large, harte2023leveraging}. Typically, item information and user history are organized into text prompts, and the model is asked to recommend the next item in a question-answer format. Researchers have discovered that LLMs, through simple in-context learning, already exhibit some capacity to perform recommendations. This suggests that LLMs, during pretraining, have learned to extract item relationships and capture basic collaborative signals, motivating further exploration of LLMs in recommendation tasks. The second approach involves fine-tuning pretrained language models specifically for recommendation, providing greater capacity for such tasks~\cite{kim2024large,zhu2024collaborative}. Some frameworks~\cite{liao2024llara} still follow the traditional retrieval-then-rank paradigm, where a simple model retrieves a set of candidates, which are then formatted into a prompt for the LLM to rerank. Other researchers~\cite{lcrec, letter} have pioneered a new pattern called generative recommendation, where the retrieval and reranking stages are merged into a single step, utilizing the LLM’s strong capabilities to directly generate the next recommended item without the need for pre-retrieval or candidate sets.

\subsection{Item Tokenization Methods in Generative Recommendation}

Using large language models (LLMs) for generative recommendations has garnered increasing attention recently. These models typically adopt a sequence-to-sequence framework to predict recommended items in one step, with user history organized into a prompt. A key challenge in this pipeline is identifying and representing items effectively. A natural approach is to use text descriptions as item identifiers~\cite{he2023large}. However, item descriptions often consist of numerous words (e.g., product titles/descriptions could contain tens or hundreds of words), making it impractical to expect an LLM to generate a complete and exact item description when recommending items, potentially leading to the hallucination problem. Additionally, representing user history in this way can result in excessively long descriptions, which reduces model efficiency. Hua et al.~\cite{cid} propose using unique numerical IDs as item identifiers, such as "1001" or "1002", which are further tokenized into several text tokens (e.g., "<00>" and "<01>") to represent items. To make these numbers meaningful, SID~\cite{cid} suggests assigning adjacent IDs to items that appear together in the same user interaction history. However, the information and relationships that simple numerical tokens can represent are limited. Moreover, there is a significant semantic gap between these numerical tokens in the LLM pretraining process and the actual meaning of the items.

To address this, some researchers have suggested using out-of-vocabulary (OOV) tokens to represent items. However, given the large number of items, assigning a unique token to each item results in considerable storage overhead. Inspired by tokenization in natural language processing, many works have attempted to represent items with a sequence of tokens, as tokens can be shared across items, reducing the number of tokens needed. SemID~\cite{cid} proposes tokenizing items using their metadata, such as item categories, encouraging items with similar metadata to share tokens. Collaborative Indexing (CID)~\cite{cid} employs hierarchical spectral clustering on the item co-occurrence matrix, where frequently co-occurring items are considered more similar and share more overlapping tokens. TIGER~\cite{tiger} introduces the use of RQVAE~\cite{rqvae} to quantize the encoded item content embeddings into tuples of semantic codewords, which can reconstruct the original content embeddings. These codewords serve as item tokens, with each code corresponding to a learned new token embedding for the LLM. Subsequent works~\cite{letter, tokenrec, wang2024enhanced} have expanded upon TIGER by integrating more collaborative information, allowing items with similar collaborative signals to share more tokens. However, these methods all rely on another model to perform the item tokenization, which may not fully align with the LLM itself. Our paper proposes using the LLM itself to refine the item tokenization process, ensuring better integration between the item representations and the LLM.
\section{Conclusion}

In this paper, we have introduced a novel approach, Self-Improving Item Tokenization (\textbf{SIIT}), which addresses key limitations in current item tokenization methods for generative recommendation systems. By allowing the large language model (LLM) to self-tune item identifiers throughout the training process, our approach mitigates inconsistencies that arise when using external models for tokenization. This self-improvement mechanism not only aligns item tokenization with the LLM's internal representations but also enhances the model's ability to capture item similarities effectively. Through extensive experimentation on diverse datasets and various initialization methods, we demonstrated that SIIT consistently improves the performance of generative recommendation systems compared to current item tokenization methods. The results underscore the importance of refining item tokenization within the generative recommendation systems, rather than relying solely on fixed, externally generated tokens. Our proposed method is simple to implement and adaptable as a plug-and-play enhancement, making it a valuable tool for researchers working with LLM-based generative recommendation systems. 
Future work could explore extending this approach to other domains where new item tokenization is critical, potentially broadening the applicability of LLMs across new domains.

\bibliographystyle{ACM-Reference-Format}
\bibliography{citation}

\appendix
\newpage
\onecolumn
\section{LETTER}\label{app:RQVAE}
LETTER add collaborative regularization and diversity regularization to the RQVAE training process. These regularizations ensure that items with similar collaborative signals have similar identifiers while maintaining diversity among the identifiers.

For the collaborative regularization, they utilize a well-trained collaborative filtering (CF) recommender model, such as SASRec, to obtain CF embeddings $h_i$ for items. Then align the quantized embedding $\hat{z}i = \sum{l=1}^L e_{t_l}$ using a contrastive loss:
\begin{align}
    \mathcal{L}_{CF} = -\frac{1}{B}\sum_{i=1}^B \frac{\text{exp}(<\hat{z}_i, h_i>)}{\sum_{j=1}^B \text{exp}(<\hat{z}_j, h_j>)}
\end{align}
where $<\cdot, \cdot>$ denotes the inner  product, and $B$ is the batch size.

The diversity regularization encourages a more uniform assignment of the code. For each cluster, they group the code center embeddings into $K$ clusters using constrained K-means and then regularize the clustered code embeddings by:
\begin{align}
    \mathcal{L}_{Div} =  -\frac{1}{B}\sum_{i=1}^B \frac{\text{exp}(<e_i^l, e^+>)}{\sum_{j=1}^B \text{exp}(<e_i^l, e_j^l>)}
\end{align}
where $e_i^l$ represents the code embedding of item $i$ at level $l$, and $e^+$ denotes the code embedding of a randomly selected sample from the same code cluster.
\section{Training and Inference Prompts Used in Sequential Recommendation Task}\label{app:prompt}

\tcbset{colframe=blue!20!gray, colback=blue!5!white, fonttitle=\bfseries}

\begin{tcolorbox}[title=Prompt for Instruments and Beauty]
1. This person has bought [HistoryHere] in the previous. Please predict the next product this person will buy. Answer: \\

2. This person has bought [HistoryHere] in the previous. Recommend one product for this person to buy next. The product you recommend is: \\

3. The shopping history of this person is: [HistoryHere]. Recommend a next product for this person to buy. Recommendation:\\

4. Based on the person's previous purchases of [HistoryHere], which product do you think he/she will buy next?  Answer:\\

5. The person has recently bought [HistoryHere]. Can you predict the next product they will buy?  Answer:\\

6. After buying [HistoryHere], which product do you think the person will choose next?  Answer:\\

7. Having previously bought [HistoryHere], which product do you anticipate the person will buy next?  Answer:\\

8. Can you predict the next product this person will buy based on their previous choice of [HistoryHere]?  Answer:\\

9. The person's shopping history includes [HistoryHere]. Now, which product do you expect him/her to buy next?  Answer:\\

10. Given that the person has already bught [HistoryHere], which product would you recommend this person to buy next?  Answer:\\

11. After buying [HistoryHere]. Can you predict her/his next choice?  Answer:\\

12. The person's shopping history contains [HistoryHere]. Please suggest the next product this person is likely to buy.  Answer:\\

13. Given the previous product choice of [HistoryHere], which product do you anticipate the person will buy next? Answer:\\

14. The person has recently bought [HistoryHere]. Now, which product do you predict this person will buy next?  Answer:\\

15. Based on the person's previous choice of [HistoryHere], which product would be the most likely next selection?  Answer:
\end{tcolorbox}

\begin{tcolorbox}[title=Prompt for Yelp]

1. This person has visited [HistoryHere] in the previous. Please predict the next restaurant this person will visit. Answer:\\

2. This person has visited [HistoryHere] in the previous. Recommend one restaurant for this person to visit next. The restaurant you recommend is:\\

3. The dining history of this person is [HistoryHere]. Recommend a next restaurant for this person to visit. Recommendation:\\

4. Based on the person's previous visits to [HistoryHere], which restaurant do you think he/she will visit next? Answer:\\

5. The person has recently visited [HistoryHere]. Can you predict the next restaurant they will visit? Answer:\\

6. After dining at [HistoryHere], which restaurant do you think the person will choose next? Answer:\\

7. Having previously visited [HistoryHere], which restaurant do you anticipate the person will visit next? Answer:\\

8. Can you predict the next restaurant this person will visit based on their previous choice of [HistoryHere]? Answer:\\

9. The person's dining history includes [HistoryHere]. Now, which restaurant do you expect him/her to visit next? Answer:\\

10. Given that the person has already visited [HistoryHere], which restaurant would you recommend this person to visit next? Answer:\\

11. After visiting [HistoryHere], can you predict her/his next choice? Answer:\\

12. The person's dining history contains [HistoryHere]. Please suggest the next restaurant this person is likely to visit. Answer:\\

13. Given the previous restaurant choice of [HistoryHere], which restaurant do you anticipate the person will visit next? Answer:\\

14. The person has recently visited [HistoryHere]. Now, which restaurant do you predict this person will visit next? Answer:\\

15. Based on the person's previous choice of [HistoryHere], which restaurant would be the most likely next selection? Answer:

\end{tcolorbox}


\section{Evaluating the Generation Variety of SIIT}\label{app:variety}

We also evaluate the generation variety of SIIT using the \textit{coverage@k} metric, which is computed as 
$Converge@k=\frac{\cup_i^N \{\text{predicted topk item in sequence i}\}}{|\text{\#item}|}$, and we also present the coverage rate of the ground truth, 
 $Converge@GT=\frac{\cup_i^N \{\text{next item in sequence i}\}}{|\text{\#item}|}$ as a reference. Our results indicate that the identifier refinement process improves the model’s ability to predict diverse items, leading to better variety. This improvement is likely due to the refined identifiers’ enhanced semantic distinctiveness, which helps the model differentiate between items with varying semantic meanings and reduces confusion.

\begin{table*}[h]
\centering
\caption{Coverage@k on Instruments, Beauty and Yelp}\label{tab:coverage}
\begin{tabular}{c|cc|cc|cc}
\toprule
\multirow{2}{*}{\textbf{Methods}} & \multicolumn{2}{c|}{\textbf{Instruments}} & \multicolumn{2}{c|}{\textbf{Beauty}} & \multicolumn{2}{c}{\textbf{Yelp}} \\ 
~ & \multicolumn{2}{c|}{\textbf{Coverage@GT=0.7131}} & \multicolumn{2}{c|}{\textbf{Coverage@GT=0.6593}} & \multicolumn{2}{c}{\textbf{Coverage@GT=0.6452}} \\ 
\textbf{Metric} & \textbf{Coverage@5} & \textbf{Coverage@10} & \textbf{Coverage@5} & \textbf{Coverage@10} & \textbf{Coverage@5} & \textbf{Coverage@10} \\
\toprule
CID & 0.3245 & 0.4767 & 0.6218 & 0.7702 & 0.5295 & 0.6980 \\
\textbf{CID+SIIT} & 0.5294 & 0.6854 & 0.6593 & 0.7460 & 0.6121 & 0.7485 \\
\midrule
TIGER & 0.0998 & 0.1484 & 0.2867 & 0.3879 & 0.2458 & 0.3540\\
\textbf{TIGER+SIIT} &  0.1402  & 0.2105 & 0.2969 & 0.4070 & 0.3637 & 0.4928\\
\midrule
LETTER & 0.1132 & 0.1702 &  0.4191 & 0.5426 & 0.3023 & 0.4112\\
\textbf{LETTER+SIIT} & 0.1743 & 0.2485 & 0.4476 & 0.5714 & 0.3185 & 0.4307 \\

\bottomrule
\end{tabular}
\end{table*}

\end{document}